# Lightweight CycleGAN Models for Cross-Modality Image Transformation and Experimental Quality Assessment in Fluorescence Microscopy

MOHAMMAD SOLTANINEZHAD[1,2], YASHAR ROUZBAHANI[3,4], JHONATAN CONTRERAS[1,2], ROHAN CHIPPALKATTI[5], DANIEL KWAKU ABANKWA[5], CHRISTIAN EGGELING[3,4], THOMAS BOCKLITZ[1,2]

[1]*Department "Photonic Data Science ", Leibniz Institute of Photonic Technology, Member of Leibniz Health Technologies, Member of the Leibniz Centre for Photonics in Infection Research (LPI), Jena, Germany*
[2]*Work group "Photonic Data Science ", Institute of Physical Chemistry (IPC) and Abbe Center of Photonics (ACP), Friedrich Schiller University Jena, Member of the Leibniz Centre for Photonics in Infection Research (LPI), Jena, Germany*
[3]*Institute of Applied Optics and Biophysics Friedrich Schiller University Jena, Jena, Germany.*
[4]*Leibniz Institute of Photonic Technologies Department of Biophysical Imaging, Jena, Germany.*
[5] *Cancer Cell Biology and Drug Discovery group, 2 Bioinformatics Core, Department of Life Sciences and Medicine, University of Luxembourg, L-4367 Esch-sur-Alzette, Luxembourg*
thomas.bocklitz@uni-jena.de

**Abstract:** With the growing integration of artificial intelligence in scientific and medical applications, lightweight deep learning models have become increasingly important. These models offer substantial reductions in memory usage and computational time. Given that GPU-based model training and inference contribute significantly to carbon emissions, lightweight architectures with comparable performance to parameter-rich models present a more environmentally friendly alternative. In this study, we focus on lightweight CycleGAN models for modality transfer in fluorescence microscopy, specifically from standard confocal to super-resolution STED and deconvolved STED microscopy images. Obtaining paired datasets in medical imaging and super-resolution microscopy is often infeasible due to the need for additional experiments and the intrinsic complexity of biological sample preparation. To address this, we investigate the performance of lightweight CycleGAN models, demonstrating their ability to achieve high-fidelity modality transfer despite reduced model complexity. We introduce a fixed channel strategy within the U-Net-based generator, in contrast to the traditional channel-doubling approach. This modification significantly reduces the number of trainable parameters—from 41.8 million to approximately nine thousand, while providing superior performance. Benefits include faster training, lower memory requirements, and reduced risk of overfitting. Furthermore, we demonstrate the utility of GAN models as diagnostic tools for experimental and labeling quality. When trained on high-quality STED or deconvolved STED microscopy images, the GAN implicitly learns the characteristics of optimal imaging. Deviations between GAN-generated outputs trained on high-quality data and low-quality experimental images can reveal issues such as photobleaching, experimental artifacts, or inaccurate labeling. This enables the model to serve as a practical tool for validating experimental accuracy and image fidelity in fluorescence microscopy workflows.

## 1. Introduction

Fluorescence microscopy is a very essential analysis tool in many biomedical and life science applications. Here, confocal microscopy provides enhanced imaging contrast over traditional light microscopy, allowing detailed imaging of cellular structures. However, its resolution is limited by the diffraction barrier, typically around 200 nm. Super-resolution microscopy, such as Stimulated Emission Depletion (STED) microscopy, surpasses this limitation by reversibly switching on and off the fluorescence emission. In STED microscopy, a depletion laser is employed to selectively deplete the fluorescence emission of fluorophores at the focal periphery, achieving resolutions beyond the diffraction limit. Routinely, STED microscopes reach a standard resolution of approximately 30-60 nm in biological samples [1, 2]. While STED microscopy offers superior spatial resolution, it presents experimental challenges such as photobleaching, phototoxicity, fluorescent labeling limitations, and experimental complexity. High-intensity depletion lasers in STED can lead to irreversible fluorophore degradation and cellular damage, limiting long-term imaging capabilities. Moreover, the intricate setup and precise alignment required for STED increases its experimental demands. These limitations motivate the pursuit of computational approaches capable of replicating STED-level resolution without requiring direct super-resolution acquisition [3, 4].

An often employed straightforward approach to potentially enhance spatial resolution in microscope images is deconvolution. Here, the image is deconvolved with the imaging or point-spread-function (PSF) of the microscope to reduce noise and enhance high spatial frequencies in the image. However, deconvolution comes with its own set of challenges and problems. Deconvolution relies heavily on high-quality microscopy images as input, as poor initial image quality can lead to errors in the final reconstructed image. One major challenge is noise amplification, where the deconvolution process enhances noise rather than resolving finer structural details, particularly in regions with low signal. This can create artifacts that obscure the true biological structures being analyzed. Computational complexity is another concern, as high-resolution microscopy images require significant processing power and time to perform deconvolution effectively. Advanced deconvolution methods can be resource-intensive, potentially slowing down analysis pipelines. Additionally, deconvolution accuracy depends on having a well-characterized PSF for the microscope; discrepancies between the assumed and actual PSF can result in distorted reconstructions. Addressing these challenges requires careful optimization of imaging conditions, robust noise suppression techniques, and advanced computational algorithms specifically adapted for the microscopy technique in use [5]. Meanwhile, artificial intelligence (AI) methods can help generate comparable images using Generative Adversarial Networks (GANs). GANs are increasingly used in medical imaging tasks such as image reconstruction, segmentation, synthesis, and cross-modality translation [6]. They have shown notable success in other imaging modalities, including Positron Emission Tomography (PET), Magnetic Resonance Imaging (MRI), and Computed Tomography (CT) by generating realistic data, improving image quality, and reducing domain shift [7]. In the field of microscopy, GANs have been employed to enhance fluorescence image quality, resolution, and modality conversion. They are particularly effective in denoising low-SNR images, enabling imaging under low illumination to minimize photobleaching and photodamage [8-12].

Studies have used various network architectures, including GANs and convolutional encoder–decoder models, to generate high-resolution microscopy images from low-resolution inputs across multiple modalities, such as structured illumination microscopy (SIM), stimulated emission depletion microscopy (STED), single-molecule localization microscopy (SMLM), confocal microscopy, and electron microscopy (EM) [13-16]. GANs have also shown strong capabilities in cross-modality transfer. For instance, models have been trained to transform images across widefield, confocal, and STED domains [17]. Other studies demonstrated SMLM super-resolution from sparse molecule data [18, 19].

In live-cell imaging, GANs have been used to enhance spatial and temporal resolution by exploiting the continuity of image sequences [20, 21]. Their application in reconstructing 3D fluorescence images from single 2D inputs further extends their utility [22]. Despite these successes, challenges such as artifact generation and limited generalization across sample types remain. Nevertheless, these studies highlight the growing impact of GANs in advancing fluorescence and super-resolution microscopy [23].

Several GAN architectures have been applied in cross-modality image transformation. These include conditional GANs (cGANs) like pix2pix, CycleGAN, and UNIT, and hybrid architectures like VAEGAN, which can be extended to conditional variants. Conditional GANs guide the generator or discriminator with additional information, such as class annotations or images, enabling more controlled generation and improving image fidelity. VAEGANs combine variational autoencoders and GANs, while UNIT and CycleGAN support unpaired data. Pix2Pix requires paired training images [24-27].

Among these, Pix2Pix and CycleGAN are widely favored [28]. Pix2Pix uses a paired learning framework, while CycleGAN enables training via cycle-consistency loss, making it suitable for unpaired learning when aligned datasets are unavailable [29]. The cycle-consistency loss enforces that an image translated from one domain to another and then back should closely match the original. This is achieved using an inverse mapping that extracts information from both modalities, enhancing learning stability. Pix2Pix is simpler and less computationally intensive; CycleGAN, though more complex, offers greater flexibility, especially valuable in medical imaging and microscopy, where obtaining paired images is often challenging. It is important to note that a GAN consists of two core components: a generator, which attempts to create realistic images, and a discriminator, which aims to distinguish between real and generated images. These two networks are trained adversarially to improve each other's performance over time. Generator networks like U-Net and ResNet are commonly used [27, 30]. U-Net offers spatial detail preservation due to its skip connections and has demonstrated strong performance in modality transfer and super-resolution tasks [31, 32].

In this study, first, we directly compare Pix2Pix and CycleGAN on our co-registered paired dataset to assess the performance difference between paired and unpaired frameworks under ideal conditions. We further evaluate the CycleGAN model in a second scenario, using nine different GAN architectures ranging from heavy to lightweight models to investigate their performance and robustness under experimental variability. While AI methods are becoming increasingly widespread across scientific and industrial domains, the need for lightweight models has become essential. As model sizes grow exponentially, so do the demands on computation, memory, and energy. Large models often require significant hardware infrastructure, making them unsuitable for resource-constrained environments. It is highly beneficial if lightweight architecture can maintain competitive accuracy, as it dramatically reduces the number of parameters and operations [33]. Reducing model size not only accelerates training time and reduces latency but also lowers memory usage, which is crucial for real-time applications and edge devices [34]. More importantly, the environmental cost of training large-scale models is substantial. Recent studies have shown that training state-of-the-art deep networks can emit several tons of $CO_2$, underscoring the carbon footprint of AI at scale [35, 36].

In our research, we focus on cross-modality image translation from confocal to STED and deconvolved STED microscopy. We emphasize the use of lightweight GAN models based on U-Net architectures, with a focus on varying channel configuration strategies within the network. One of the key challenges in GAN training is achieving a balance between the generator and discriminator [26]. In this study, we compare nine different models, introducing a U-Net-based fixed channel strategy. The fixed channel strategy contributes to preserving key structural details while requiring a relatively small number of parameters, reducing the risk of artifacts introduced by deeper or variable-capacity models. Additionally, we demonstrate that, when trained on high-quality datasets, these models can also serve as tools to evaluate the quality of STED microscopy experiments. We utilized our approach on confocal and STED microscopy images of ADP-ribosylation factor-like 13B(ARL13B), a ciliary membrane protein localized to the cilium, labeled with Alexa Fluor 594 for fluorescence imaging. Ciliary proteins are of particular interest, as cilia are slender organelles with structural and functional complexities that are critical for cellular signaling and motility. High-resolution imaging of these proteins enhances our understanding of ciliary assembly, function, and associated pathologies, thereby contributing to the development of targeted therapeutic strategies [37].

## 2. Methods

*2.1 Biological Sample Preparation and Imaging Acquisition*

The dataset used in this study was derived from fluorescence microscopy imaging of ARL13B-labeled primary cilia. ARL13B is a small Ras-family GTPase anchored to the ciliary membrane and widely used as a marker for cilia structure and function. For confocal and STED microscopy, cilia were fluorescently labeled using a rabbit polyclonal anti-ARL13B primary antibody (177-11-1AP, Proteintech; 1:500), followed by a goat anti-rabbit Alexa Fluor 594 secondary antibody (A-11012, Thermo Fisher Scientific; 1:200), according to standard immunofluorescence protocols.

STED and confocal microscopy images were acquired on an Abberior Infinity Line microscope (Abberior GmbH, Göttingen, Germany) equipped with Imspector software version 16.3.16118-w2224 and an Olympus 60 × oil objective (1.42 NA), and using a 561 nm picosecond pulsed laser diodes, a 775 nm pulsed STED laser (NKT Photonics Switzerland GmbH, Regensdorf, Switzerland) and an avalanche photodiode detector (Excelitas Technologies Corp., USA) with 5 µs dwell time and line averaging of 10. A total of 256 high-quality confocal and STED images, 166 high-quality deconvolved STED images from fully co-registered image sets were acquired. Each set consists of confocal, STED, and deconvolved STED images obtained from the same field of view. Deconvolved STED images were generated from raw STED microscopy data using the Classic Maximum Likelihood Estimation (CMLE) algorithm implemented in Huygens Professional software (Scientific Volume Imaging, The Netherlands). The CMLE algorithm improves spatial resolution by iterative deconvolution based on the system point spread. In addition to the high-quality dataset, an independent set of low-quality STED and corresponding deconvolved STED images was acquired under sub-optimal imaging conditions to evaluate model robustness. This low-quality STED dataset was not used for training; the confocal image pairs served as a test set. All datasets were verified for spatial co-registration and used for the deep learning experiments (see Figure 1).

*2.2 Image processing and Augmentation:*

Prior to training, microscopy images required systematic preprocessing to ensure compatibility with the neural network input and to enhance image quality for model learning (see Figure 2a). To ensure pixel-wise correspondence between modalities, we first performed rigid coregistration between confocal and STED images using a mutual information–based histogram matching approach [38]. Next, image contrast and brightness were enhanced using ImageJ', which applies a histogram-based contrast stretching to improve signal-to-noise ratio [39]. Segmentation was then performed to isolate individual cilia structures and crop each field of view around the target region.

To match the input requirements of CNN-based architecture and to maintain image spatial resolution, the cropped images were padded to achieve a standardized dimension of 128 × 128 pixels. This size was chosen because CNN architectures benefit from input dimensions that are powers of two to optimize computational efficiency, particularly on GPU hardware, and to facilitate consistent down-sampling and up-sampling operations [40].

Before data augmentation, normalization was applied to standardize image intensity distributions across the dataset. For data augmentation, we applied only conservative geometric transformations to preserve biological structures. (Figure 2a) Specifically, random rotations (90°, 180°, 270°) and horizontal and vertical flipping were used to increase dataset variability while retaining the complete structural information of the cilium. No additional intensity or scale augmentations were applied to maintain the integrity of fluorescence signal distributions across the dataset.

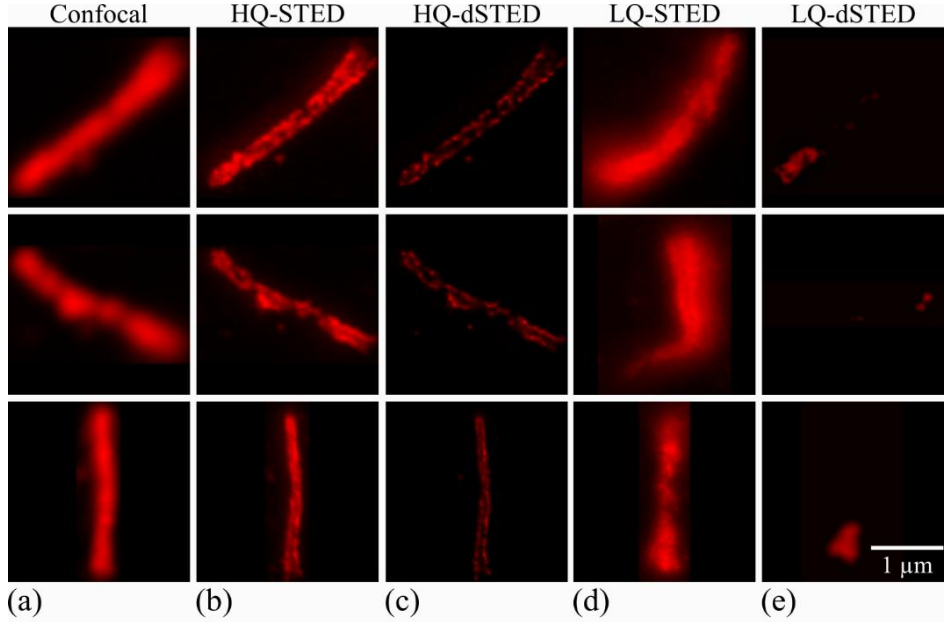

Figure 1. Representative datasets overview of confocal and differently processed images of immunolabeled ARL13B at primary cilia cells. Each row of panel columns a-c and of d-e presents a different sample: (a) confocal, (b) corresponding STED, (c) deconvolved STED, (d) low-quality STED, and (e) deconvolved low-quality STED images. All images feature the same spatial scale.

*2.3 GPU Utilization, Reinitialization, and Cross-Validation in Model Training*

We utilized an NVIDIA GeForce RTX 4090 GPU for training and evaluating our deep learning models. Built on the Ada Lovelace architecture. Its architecture supports 16,384 CUDA cores and advanced Tensor Cores, making it particularly well-suited for computationally intensive tasks such as training Generative Adversarial Networks (GANs) and handling large-scale image datasets [41].

To ensure reproducibility and prevent unintended retention of model states across training sessions, we implemented a reinitialization protocol prior to each training run. This protocol involved clearing the GPU memory cache, resetting random seeds across Python libraries, and invoking garbage collection routines. Additionally, any existing model checkpoints were deleted to guarantee a fresh start. Neglecting this reinitialization step can lead to the persistence of learned weights and random states, potentially causing data leakage and compromising the integrity of cross-validation procedures.

Our training regimen incorporated a five-fold cross-validation strategy, wherein the dataset was partitioned into five subsets. In each iteration, four subsets were used for training while the remaining one served as the validation set. This approach allowed us to assess the model's generalization capabilities and ensure that performance metrics were not biased by any particular data split.

The combination of the RTX 4090's computational prowess, rigorous reinitialization protocols, and robust cross-validation techniques contributed to the development of reliable and reproducible results in our research.

*2.4 GAN Architectures and Training Strategy*

We compared two generative adversarial network (GAN) architectures: Pix2Pix and CycleGAN. The generator network for all models was based on the U-Net architecture. The discriminator used a PatchGAN configuration to evaluate local image patches and discriminate between real and fake images [42].

    Pix2Pix was applied with paired confocal and STED/deconvolved STED images. It minimizes a combination of L1 loss (Equation (2)), to encourage similarity between generated and target images, and adversarial loss from the discriminator (Equation (3)). The full Pix2Pix objective is given in Equation (1) [27].

$$\text{Loss}_{\text{Pix2Pix}}(G, D) = \text{Loss}_{\text{GAN}}(G, D) + \lambda \cdot \text{Loss}_{L1}(G) \quad (1)$$

$$Loss_{L1}(G) = E_{x,y}[||y - G(x)||_1] \quad (2)$$

$$\text{Loss}_{\text{cGAN}}(G, D) = E_{x,y}[\log D(x, y)] + E_x\left[\log\left(1 - D(x, G(x))\right)\right] \quad (3)$$

Here, G and D denote the generator and discriminator networks, respectively, while x represents the input confocal image and y is the corresponding STED or deconvolved STED image. The symbol E represents the expectation over the training data, and $\|\cdot\|_1$ indicates the L1 norm. The scalar λ balances the contribution of the adversarial loss and the L1 reconstruction loss.

For Pix2Pix, only the most complex generator architecture was used as a reference, whereas the full channel reduction study was conducted exclusively for CycleGAN models.

CycleGAN employs two coupled generator–discriminator networks to perform unpaired image-to-image translation. Its objective function combines an adversarial loss, a cycle-consistency loss to enforce reversibility between source and target domains, and an optional identity loss to encourage structural and intensity preservation when the source and target domains are similar. The total loss is defined in Equation (4), with the cycle-consistency and identity components detailed in Equations (5) and (6), respectively [28].

$$\text{Loss}_{\text{CycleGAN}} = \text{Loss}_{\text{GAN}}(G, D_Y) + \text{Loss}_{\text{GAN}}(F, D_X) + \lambda_{\text{cyc}} \cdot \text{Loss}_{\text{cyc}}(G, F) + \lambda_{\text{id}} \cdot \text{Loss}_{\text{id}}(G, F) \quad (4)$$

$$\text{Loss}_{\text{cyc}}(G, F) = E_x[|F(G(x)) - x|_1] + E_y[|G(F(y)) - y|_1] \quad (5)$$

$$\text{Loss}_{\text{id}}(G, F) = E_y[|G(y) - y|_1] + E_x[|F(x) - x|_1] \quad (6)$$

In these equations, F denotes the inverse generator mapping from the target domain to the source. $D_X$ and $D_Y$ are the discriminators for the source and target domains, respectively. The scalar $\lambda_{\text{cyc}}$ controls the importance of the cycle-consistency loss, while $\lambda_{\text{id}}$ weights the identity loss. The term $\text{Loss}_{\text{cyc}}$ ensures that mappings are reversible (i.e., x→y→x), and $\text{Loss}_{\text{id}}$ encourages generators to behave like identity mappings when source and target domains are aligned, such as in STED → STED translation.

To investigate the influence of model complexity on performance, we designed a series of generator variants with systematically reduced numbers of feature channels. Two distinct architectural strategies were implemented:

Doubling channel architecture: The number of feature channels doubled after each down-sampling layer, following the classical U-Net design. Across models, the initial number of feature channels was progressively halved — for example, starting with 64 feature channels and a bottleneck of 512 channels (Model 1), down to 4 channels with a 64-channel bottleneck (Model 6).

Fixed channel architecture: Each layer used a fixed number of feature channels across the entire network. This number was progressively reduced across models — from 64 feature channels per layer (Model 6) down to 8 feature channels per layer (Model 9).

The model index increases as the architectural complexity decreases. All models were trained on the co-registered confocal and STED dataset under identical conditions to ensure consistency across experiments (Figure 2b, c).

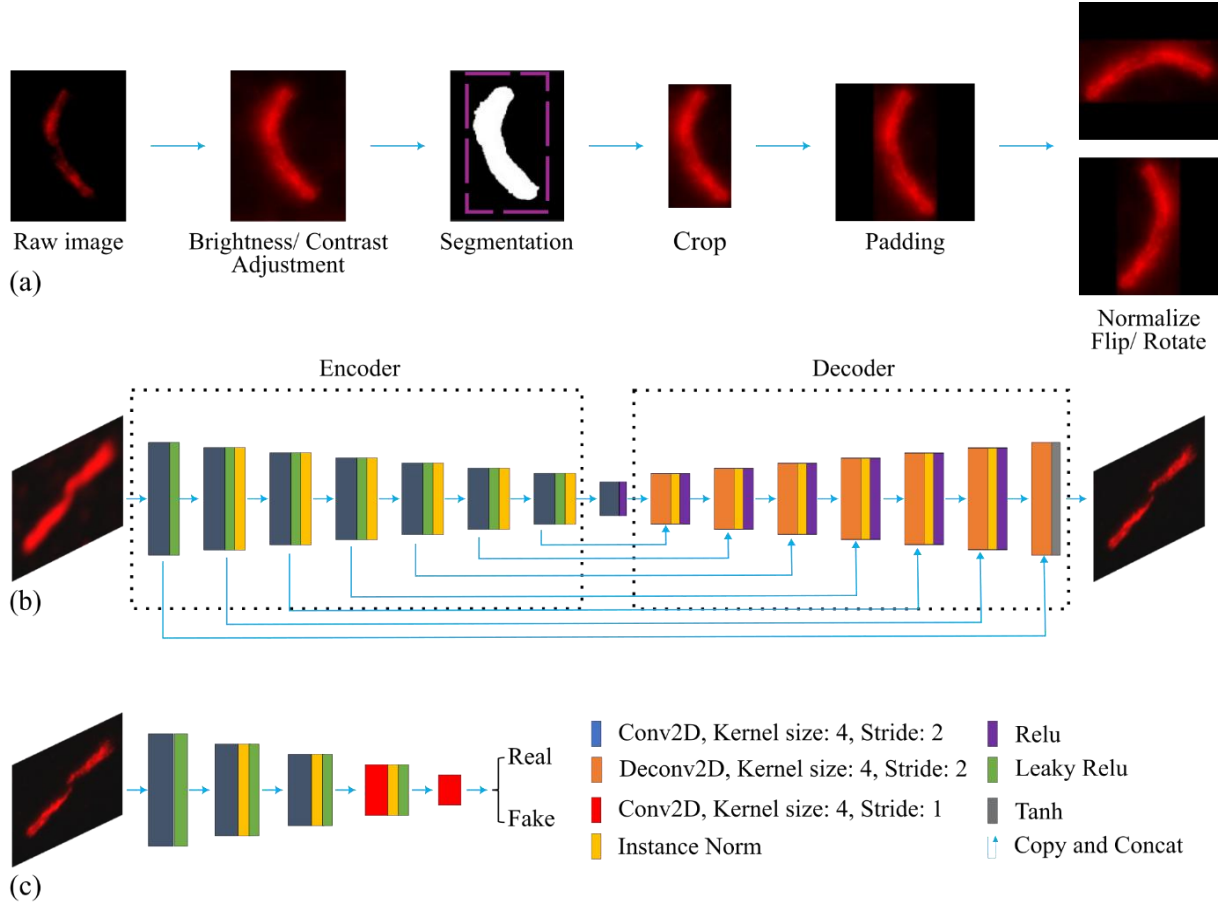

Figure 2. Image preprocessing workflow and GAN model architecture.
(a) Schematic illustration of the image preprocessing and data augmentation pipeline. The workflow begins with brightness and contrast enhancement using intensity histogram normalization to standardize image brightness and contrast across samples. This is followed by segmentation and cropping of the cilium region of interest, then padding to a fixed size of 128 × 128 pixels. Finally, images are normalized and augmented through random rotations (in 90° increments) and horizontal/vertical flipping to improve model generalization.
(b, c) Architectures of the generator and discriminator networks. The generator is a U-Net with a symmetric encoder-decoder structure and skip connections, enabling preservation of spatial features across scales. The discriminator follows the PatchGAN design, evaluating overlapping image patches to distinguish real from generated images, thereby encouraging the generator to produce realistic local textures.

## 2.5 Evaluation Metrics

To quantitatively assess the quality of the generated images, we used two established image similarity metrics: Peak Signal-to-Noise Ratio (PSNR) and Structural Similarity Index Measure (SSIM) [43,44,45].
PSNR measures the logarithmic ratio between the maximum possible pixel value, and the mean squared error (MSE) between the reference and test image. Higher PSNR values indicate lower error and better reconstruction quality. SSIM was developed to better align with human visual perception. It evaluates local patterns of luminance, contrast, and structural similarity between images, providing a value between -1 and 1, with 1 indicating perfect similarity, 0 indicating no similarity, and -1 indicating perfect anti-correlation.
All metrics were computed on the full dataset by aggregating results from each fold of the 5-fold cross-validation, such that every image served as part of the validation set in at least one fold. This allowed us to compare model outputs against the ground truth STED and deconvolved STED images. These quantitative metrics were complemented by qualitative inspection to assess biological relevance and fine structural details (see Equations 7,8,9).

$$\text{PSNR} = 10 \cdot \log_{10}\left(\frac{\text{MAX}_I^2}{\text{MSE}}\right) \quad (7)$$

While MSE is

$$\text{MSE} = \frac{1}{mn}\sum_{i=0}^{m-1}\sum_{j=0}^{n-1}[I(i,j) - K(i,j)]^2 \quad (8)$$

$$\text{SSIM}(x,y) = \frac{(2\mu_x\mu_y + C_1)(2\sigma_{xy} + C_2)}{(\mu_x^2 + \mu_y^2 + C_1)(\sigma_x^2 + \sigma_y^2 + C_2)} \quad (9)$$

In Equation (7), $MAX_I$ denotes the maximum possible pixel value of the image, and MSE is defined in Equation (8) as the average squared pixel-wise difference between the reference image I and the generated image K, over image dimensions m×n. In Equation (9), $\mu_x$ and $\mu_y$ represent the local mean intensities of patches x and y, $\sigma_x^2$ and $\sigma_y^2$ their variances, and $\sigma_{xy}$ the covariance. Constants C1 and C2 are included to avoid instability in regions with low denominator values.

## 3. Results and Discussion

### 3.1 Pix2Pix vs. CycleGAN

Using differently trained algorithms, we aimed at cross-modality image translation from confocal to STED and deconvolved STED microscopy images. As an experimental example, we took multiple sets of confocal and STED microscopy images of immunolabeled ARL13B at primary cilia cells. We started with CycleGAN and Pix2Pix algorithms.

We first evaluated the performance of CycleGAN and Pix2Pix using the paired confocal and STED microscopy images. With paired data, we have known matches from both modalities as ground truth, which is not available in the unpaired scenario. Using paired data to test CycleGAN is an effective strategy to assess its performance in settings where ground truth is otherwise unavailable. In our experiment, Pix2Pix slightly outperformed CycleGAN. However, CycleGAN still achieved competitive results. For the confocal to STED modality transfer task, Pix2Pix achieved an average SSIM of 0.93 and a normalized PSNR of 0.61, both computed using the ground truth STED images as reference. In comparison, CycleGAN reached an SSIM of 0.90 and a normalized PSNR of 0.46, while the confocal baseline had an SSIM of 0.60 and a normalized PSNR of 0.27 (Figure 3b).

Figure 3a presents an example of the confocal-to-STED modality transfer task. From left to right, the images show the input confocal image, CycleGAN-generated STED, Pix2Pix-generated STED, and the ground truth (real) STED image.

In Figure 3c, the intensity profile is plotted along the yellow line shown in the images in Figure 3a. Based on the extracted intensity profiles, both Pix2Pix and CycleGAN demonstrate the ability to reconstruct structural features from the confocal input, with varying degrees of fidelity. Notably, both models recover the presence of two distinct peaks, while the confocal profile resembles a single broad Gaussian-like shape. The Pix2Pix generation closely follows the STED ground truth, with a slight tendency to overestimate peak intensities, whereas CycleGAN slightly underestimates them. In contrast, the confocal profile exhibits broader and less defined peaks, consistent with its lower resolution. These observations are supported by Pearson correlation coefficients with the ground truth profile: Pix2Pix (0.976), CycleGAN (0.927), and confocal (0.893).

For the confocal-to-deconvolved STED (dSTED) task, average SSIM and normalized PSNR values across the test dataset show Pix2Pix (SSIM: 0.89, PSNR: 0.69) outperforms CycleGAN (SSIM: 0.88, PSNR: 0.62) and the confocal input (SSIM: 0.50, PSNR: 0.17), as shown in Figure 4b.

Figure 4a shows, from left to right, the confocal input, CycleGAN output, Pix2Pix output, and the ground truth dSTED image. Both models aim to generate the corresponding dSTED modality from the confocal input. In Figure 4c, the intensity profile extracted along the yellow lines in Figure 4a illustrates the ability of the models to recover fine structural features from the confocal input. Both Pix2Pix and CycleGAN successfully reconstruct the two-peak structure and capture the overall profile shape of the ground truth. However, both models overestimate the peak intensities, with CycleGAN slightly exaggerating the second peak more prominently. The confocal input shows a broadened signal with reduced resolution. These observations are reflected in the Pearson correlation coefficients with the ground truth: Pix2Pix (0.868), CycleGAN (0.823), and confocal (0.803), highlighting the improved structural fidelity achieved by the GAN-based models.

These results highlight the comparative strengths and limitations of Pix2Pix and CycleGAN in fluorescence microscopy modality transfer tasks. While Pix2Pix consistently outperforms CycleGAN in SSIM, PSNR, and intensity profile fidelity, its dependence on paired data limits its applicability in many real-world scenarios. CycleGAN, although slightly less accurate, demonstrates strong reconstruction capabilities without requiring exact input-output alignment, making it a robust alternative when paired data are unavailable. The large performance gap between the confocal baseline and both GAN models also underscores the effectiveness of deep learning-based cross-modality translation in enhancing resolution and contrast.

### 3.2 Evaluating CycleGAN Robustness with Decreased Network Complexity

Next, we evaluated the modality transfer performance for both confocal to STED and confocal to deconvolved STED data using GAN models. The architecture is progressively simplified in complexity, with the number of parameters decreasing from 41.8 million in Model 1 to approximately 9 thousand in Model 9 (Figure 5a).

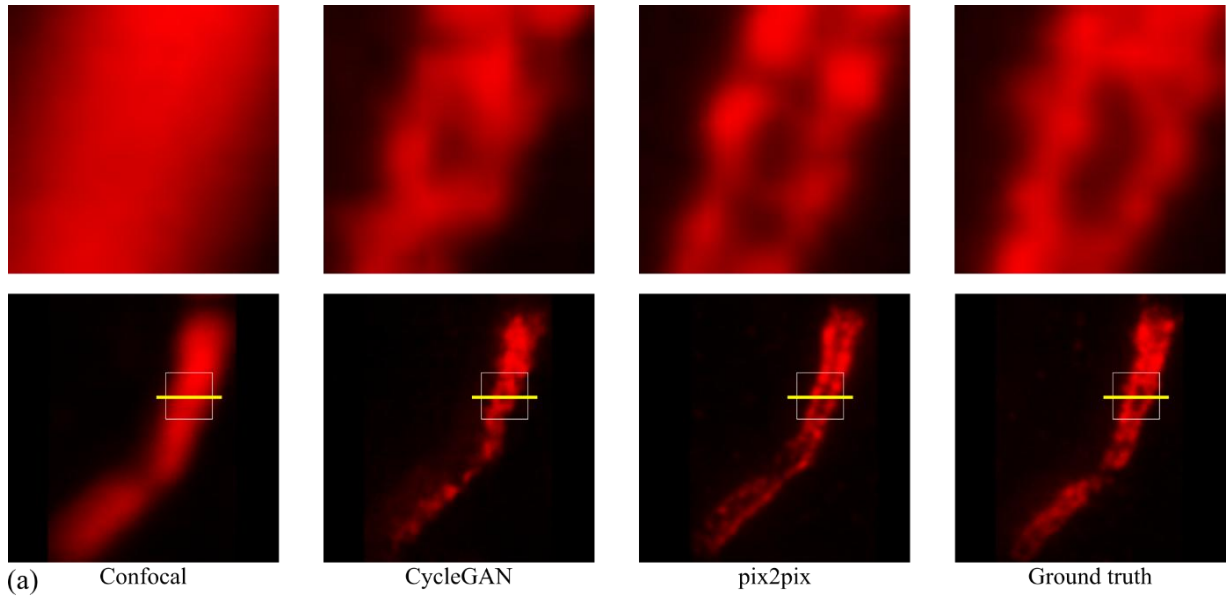

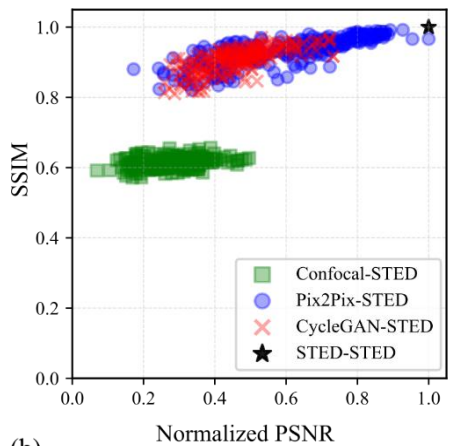
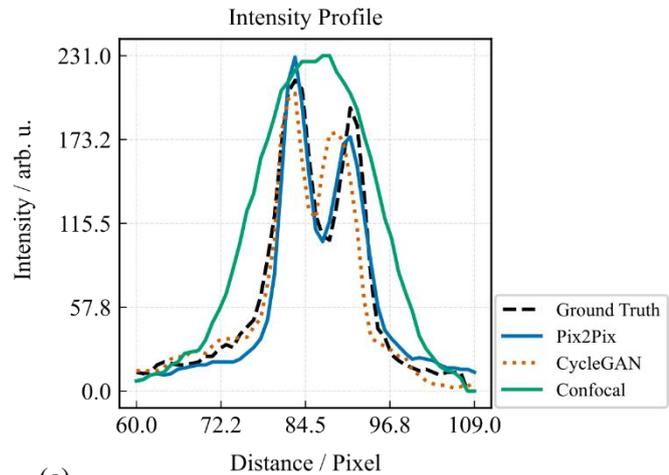

Figure 3. Comparison between Pix2Pix and CycleGAN methods for Confocal-to-STED cross modality Transfer.

(a) Representative images of the confocal to STED translation task. From left to right: input confocal image, CycleGAN-generated STED image, Pix2Pix-generated STED image, and the STED ground truth. White boxes indicate regions shown in the zoomed-in views, and yellow lines mark the positions used for intensity profile analysis.

(b) Comparison of image quality metrics (SSIM vs. normalized PSNR) for Pix2Pix-generated, CycleGAN-generated, and Confocal images. Each point represents an individual test sample, with marker color and shape indicating the method. The black star at (1.0, 1.0) denotes the STED-STED reference. Mean values: SSIM (Pix2Pix = 0.93, CycleGAN = 0.90, Confocal = 0.60); Normalized PSNR (Pix2Pix = 0.61, CycleGAN = 0.46, Confocal = 0.27).

(c) Intensity profiles along the yellow lines in (a), comparing the confocal input, CycleGAN and Pix2Pix outputs, and the ground truth STED image. Both models recover the two-peak structure, in contrast to the confocal profile, which appears as a single broad peak due to limited resolution. Pearson correlations with the ground truth: Pix2Pix (0.976), CycleGAN (0.927), confocal (0.893). All images feature the same spatial scale.

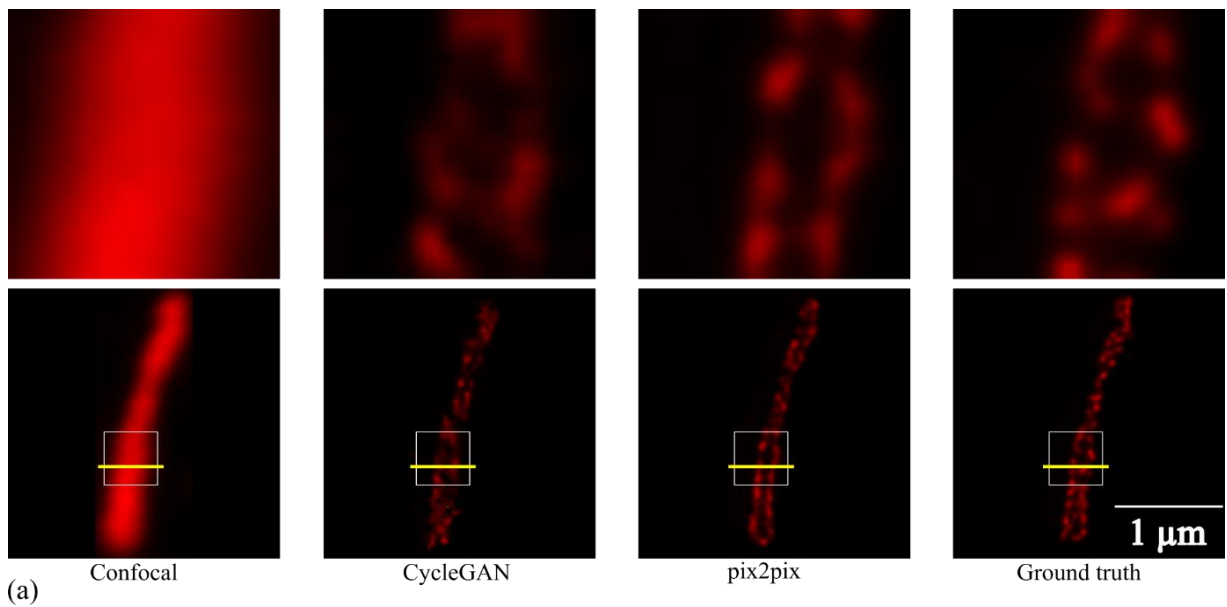

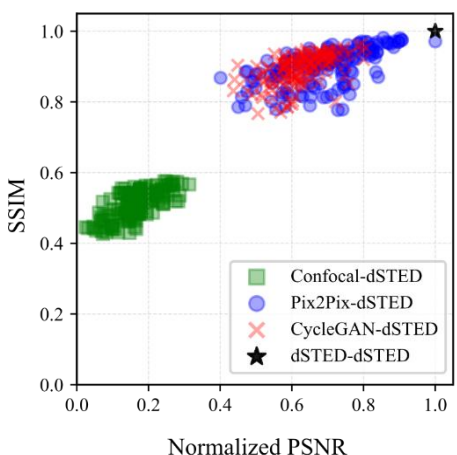

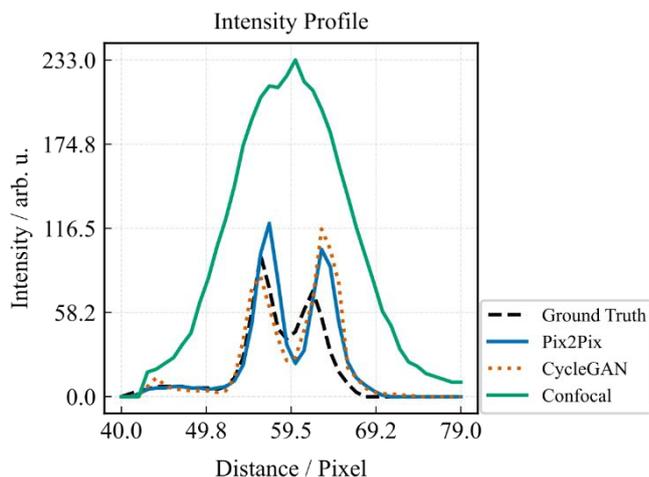

Figure 4. Comparison between Pix2Pix and CycleGAN methods for confocal to deconvolved STED (dSTED) cross modality transfer.

(a) Representative images of the confocal to dSTED translation task. From left to right: input confocal image, CycleGAN-generated dSTED image, Pix2Pix-generated dSTED image, and the dSTED ground truth. White boxes indicate regions shown in the zoomed-in views, and yellow lines mark the positions used for intensity profile analysis.

(b) Comparison of image quality metrics (SSIM vs. normalized PSNR) for Pix2Pix-generated, CycleGAN-generated, and Confocal images. Each point represents an individual test sample, with marker color and shape indicating the method. The black star at (1.0, 1.0) denotes the dSTED-dSTED reference. Mean values: SSIM (Pix2Pix = 0.89, CycleGAN = 0.88, Confocal = 0.50); Normalized PSNR (Pix2Pix = 0.69, CycleGAN = 0.62, Confocal = 0.17).

(c) Intensity profiles along the yellow lines in (a), comparing the confocal input, CycleGAN and Pix2Pix outputs, and the ground truth dSTED image. Both Pix2Pix and CycleGAN recover the two-peak structure, although CycleGAN underestimates the second peak. The confocal profile remains broad and unresolved. Pearson correlation with the ground truth: Pix2Pix (0.868), CycleGAN (0.823), confocal (0.803). All images feature the same spatial scale.

Correspondingly, the storage size of the saved weight decreases from 12.8 GB to 67 MB; all models were trained for 200 epochs, with model checkpoints saved every 5 epochs (Figure 5b).

Table 1 summarizes the generator parameter counts and storage requirements, enabling a direct comparison of model efficiency, scalability, and performance across architectures with varying capacities. The inference time was measured using the largest model, with results shown in Figure 5c and Table 2. As the number of test images increased from 1 to 256, the average inference time per image dropped from 4.89 s to 0.0255 s. This behavior is consistent with observations in GPU computing literature, where fixed overheads such as kernel launch, memory transfer, and synchronization dominate at small batch sizes but have diminishing impact at larger batch sizes [41]. Additionally, larger batches enable better parallelism and more efficient use of GPU cores, reducing idle time and maximizing throughput [46].

In the confocal-to-STED modality transfer task, image quality metrics across the nine GAN models represent averages computed over the entire test dataset and demonstrate a high degree of consistency. The highest SSIM value of 0.910 is achieved by Model 8, while the lowest is 0.878 in Model 6. For PSNR, the maximum value of 26.62 dB is observed in Model 1, and the minimum is 26.11 dB in Model 5. While the overall results are tightly clustered, slight performance differences are evident. Specifically, Models 1 and 8 exhibit slightly superior metrics in PSNR and SSIM, respectively, indicating marginally better structural and perceptual fidelity. These observations are further illustrated in Figure 6b and quantitatively summarized in Table 3, supporting the robustness of cross-modality transfer performance across varying model architectures. These observations are further illustrated in Figure 6b and quantitatively summarized in Table 3, which supports the robustness of the cross-modality transfer performance across varying model architectures.

In the confocal-to-deconvolved STED modality transfer task, image quality metrics across the nine GAN models again show strong consistency, with all values representing averages computed over the entire test dataset. SSIM ranges from a minimum of 0.866 (Model 2) to a maximum of 0.906 (Model 4). PSNR values range from 26.25 dB (Model 2) to a peak of 27.29 dB (Model 4), suggesting improved reconstruction quality in the latter. Although the metric values are generally close across models, slight performance advantages can be observed. In particular, Model 4 consistently achieves the best results across both metrics, with Models 6, 7, and 8 also performing competitively, indicating superior structural preservation and perceptual quality. These differences, though subtle, are evident in Figure 7b and are quantitatively detailed in Table 5, reinforcing the robustness and consistency of the GAN-based modality transformation pipeline.

Overall, models using the fixed channel policy, particularly Models 5, 7, 8, and 9, demonstrate that simpler, uniformly scaled architectures can perform competitively with more complex designs, especially when optimized for cross-modality transfer tasks. These models achieve strong performance with reduced computational demand. The fixed channel design offers better parameter efficiency, especially in low-data regimes, and improved training stability. Such uniformity reduces the risk of issues like vanishing or exploding gradients, which can impede training in deeper networks. By maintaining consistent gradient magnitudes, these architectures enhance training stability and convergence speed [47, 48]. These combined benefits suggest a potential architectural advantage that balances simplicity with high performance, making it a compelling strategy for GAN-based image translation in microscopy. While lighter architectures show strong performance, they remain at risk of losing fine structural details. Model 8 stands out among the fixed-channel architectures, achieving the highest SSIM in the confocal-to-STED task while maintaining competitive PSNR values. This indicates that a balanced, moderately lightweight design can provide both parameter efficiency and strong reconstruction quality. Nevertheless, subtle differences in structural detail and sharpness remain, as seen in the example outputs from all nine models in Figure 6a and Figure 7a for the confocal-to-STED and confocal-to-deconvolved STED transformations, respectively. Continuing with the confocal-to-STED example, image quality metrics were computed for a single representative sample across all models. The SSIM values range from 0.93 to 0.98, with Model 9 achieving the highest score, indicating strong structural similarity to the ground truth. PSNR varies from 28.8 dB (Model 5) to 32.1 dB (Model 7), with Models 4, 7, 8, and 9 exceeding 30 dB, reflecting improved noise suppression and intensity accuracy. These results suggest that while most models deliver high perceptual and structural quality on this sample, Model 7 stands out for the highest PSNR, and Model 9 achieves the best SSIM. Model 4 also performs consistently well across all three metrics. In contrast, Models 2 and 5 exhibit slightly lower performance, especially in PSNR. Overall, these sample-level metrics reinforce the earlier trend: models employing a fixed channel policy (e.g., 4, 7, 9) tend to offer a favorable balance between image quality and architectural simplicity. The numerical values are detailed in Table 4, corresponding to the visual outputs in Figure 6a. In Figure 6c, the intensity profile is plotted along the yellow line shown in the images in Figure 6a. Model 1 (41.8 M parameters) and Model 7 (0.13 M parameters) both reconstruct the two-peak structure from the confocal input, capturing key spatial features present in the STED ground truth. Model 7 aligns closely with the ground truth in both peak position and relative intensity, while Model 1 shows a slight underestimation of the second peak. The confocal profile remains broader, with less distinct peak separation, consistent with its lower resolution. These observations are supported by Pearson correlation coefficients with the ground truth profile: Model 7 (0.9715), Model 1 (0.9411), and confocal (0.94). For the confocal-to-deconvolved STED modality transfer example shown in Figure 7a, image quality metrics were computed for a single representative sample across all nine GAN models. SSIM values are tightly grouped, with most models scoring 0.94, except Models 4, 7, and 8, which score slightly lower at 0.93 and 0.92, indicating minimal structural variation among outputs. PSNR values range from 31.0 dB (Model 8) to 33.3 dB (Model 9), with Models 7 and 9 achieving the highest scores, reflecting superior noise reduction and intensity fidelity. While all models deliver comparably high image quality on this sample, Model 9 stands out with the highest PSNR and SSIM scores, indicating excellent performance in both structure and contrast preservation. Model 7 also performs well. On the other hand, Model 8 shows slightly lower performance in PSNR and SSIM, suggesting minor degradation in sharpness or intensity. These detailed metrics are presented in Table 6 and visually supported by the example reconstructions shown in Figure 6a, providing further insight into architectural performance on a per-sample basis.

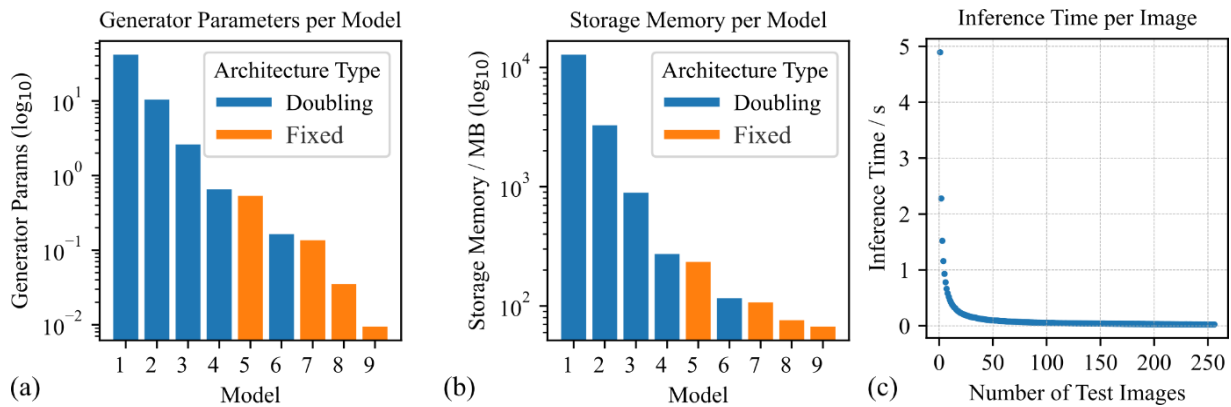

Figure 5. Summary of model complexity and performance characteristics across GAN architectures, including generator parameter count and storage memory (in megabytes). Inference time is reported for Model 1.

(a) Number of parameters in the generator networks across all nine models. Models using a doubling channel strategy are shown in blue; those using a fixed channel strategy are shown in orange. The doubling architecture starts with 64 channels and reaches a bottleneck of 512 in Model 1 (41800000 parameters), down to Model 6, which starts with 4 channels and has a bottleneck of 32. The fixed channel architecture begins with 64 channels in Model 5 and decreases to 8 channels in Model 9 (9000 parameters). For the number of parameters, see Table 1.

(b) Storage usage for different models over 200 epochs, with checkpoints saved every 5 epochs (Table 1).

(c) Inference time for Model 1, evaluated with varying numbers of test images (Table 2).

**Table 1. Model's characteristics**

| Model | 1 | 2 | 3 | 4 | 5 | 6 | 7 | 8 | 9 |
|---|---|---|---|---|---|---|---|---|---|
| Generator Parameters (M) | 41.8 | 10.45 | 2.61 | 0.65 | 0.53 | 0.16 | 0.13 | 0.035 | 0.009 |
| Storage Memory (MB) | 12800 | 3270 | 891 | 273 | 234 | 116 | 107 | 76 | 67 |

**Table 2. Inference time per image**

| Number of images | 1 | 25 | 50 | 100 | 150 | 200 | 250 |
|---|---|---|---|---|---|---|---|
| Time (Seconds) | 4.89 | 0.19 | 0.097 | 0.053 | 0.038 | 0.030 | 0.025 |

In Figure 7c, the intensity profile is plotted along the yellow line shown in the images in Figure 7a. Both Model 3 and Model 9 recover the two-peak structure present in the ground truth STED image, while the confocal profile remains broad and unresolved. Model 9, despite having significantly fewer parameters (0.009 M vs. 2.61 M in Model 3), achieves slightly better alignment with the ground truth in terms of peak position and separation. These trends are reflected in the Pearson correlation coefficients with the ground truth profile: Model 9 (0.929), Model 3 (0.923), and confocal (0.731).

On average, the CycleGAN models demonstrated slightly better SSIM scores in the confocal-to-STED modality transfer, while the confocal-to-deconvolved STED task yielded higher PSNR values. For the STED dataset, across all models, the mean SSIM and PSNR values were 0.90 and 26.41 dB, respectively. In contrast, the deconvolved STED dataset showed averages of 0.88 for SSIM and 26.94 dB for PSNR. These results suggest that while STED images preserve more structural similarity and perceptual detail, deconvolved STED images provide better intensity fidelity due to improved noise suppression [49]. This difference stems from the nature of the target data: deconvolved STED images have higher signal-to-noise ratios [50, 51]. The improved image quality facilitates more stable training and better intensity accuracy. Thus, each modality presents unique challenges and advantages that influence the performance of GAN-based image translation models. CycleGAN maintains strong performance even with drastic reductions in model size, showing high robustness across both tasks. Models with fixed channel designs balance simplicity and quality, offering stable training behavior, strong convergence, and competitive image quality metrics.

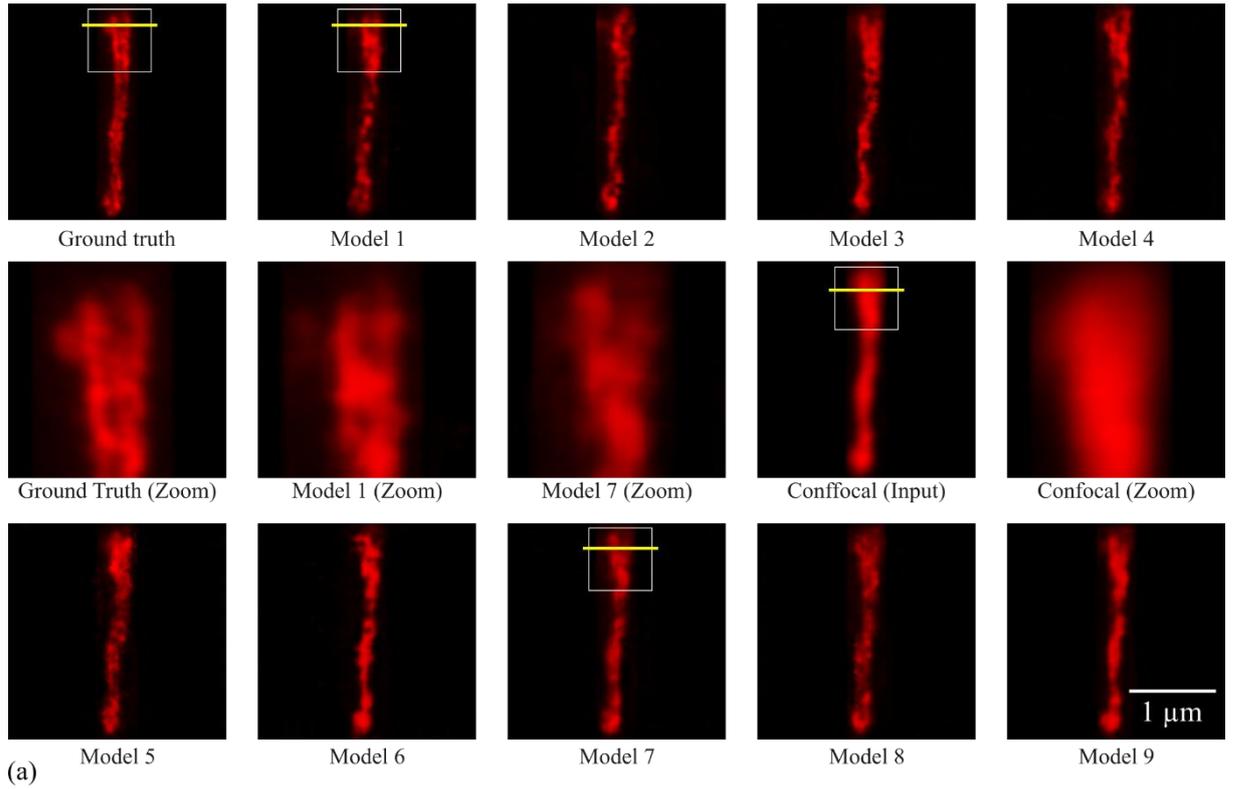

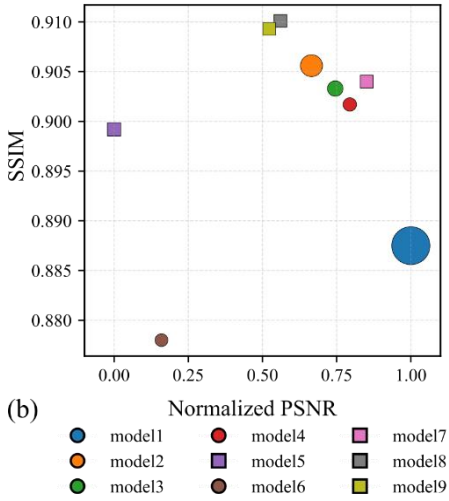
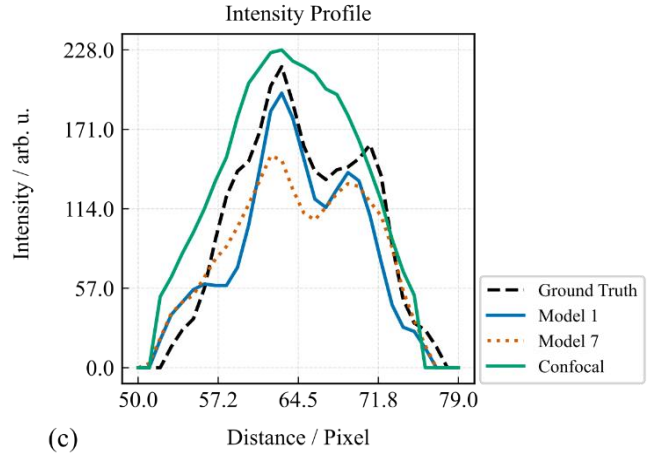

Figure 6. Quantitative and qualitative evaluation of nine CycleGAN models for confocal-to-STED modality transfer.
(a) Representative reconstruction results for a single confocal input image. The first image in the top row shows the ground truth STED image, followed by outputs from Models 1–9 in both the first and third rows. The second row shows, in order: zoomed-in crops of the ground truth, Model 1, Model 7, and the confocal input, followed by a zoomed-in view of the same region in the confocal image. This layout highlights differences in structural fidelity and signal localization. Yellow lines indicate the positions used for intensity profile analysis. PSNR and SSIM values for this sample, computed with respect to the ground truth, are reported in Table 4.
(b) Mean normalized PSNR and SSIM across the test dataset (Table 3). Each marker represents one model, with marker shape indicating the channel policy (circle: doubling; square: fixed) and marker size proportional to the number of generator parameters (see Table 1 for parameter counts).
(c) Intensity profiles plotted along the yellow lines in Figure 6a, comparing the confocal input, outputs from Model 1 and Model 7, and the ground truth STED image. Both models successfully reconstruct the two-peak structure present in the ground truth, with Model 7 aligning more closely in both peak height and position. Model 1 slightly underestimates the second peak, while the confocal profile appears broader and less resolved. These observations are supported by Pearson correlation coefficients with the ground truth profile: Model 7 (0.9715), Model 1 (0.9411), and confocal (0.94). All images feature the same spatial scale.

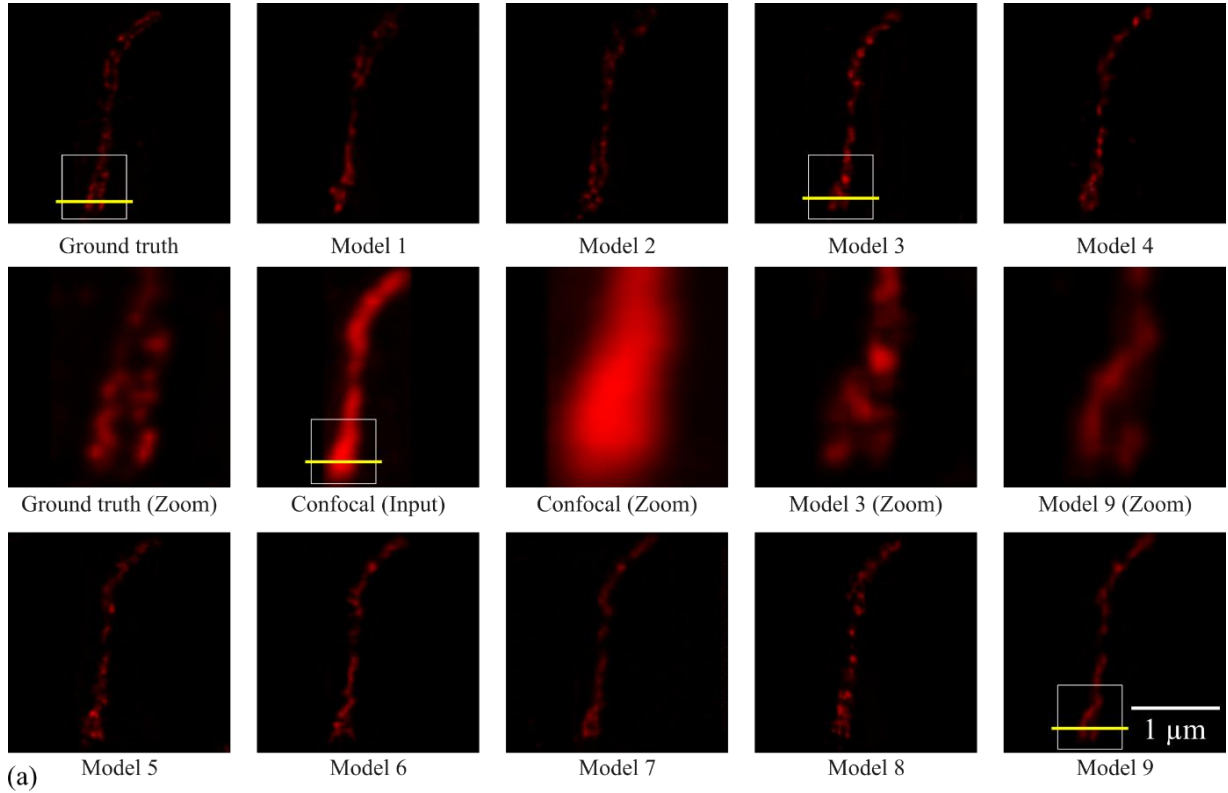

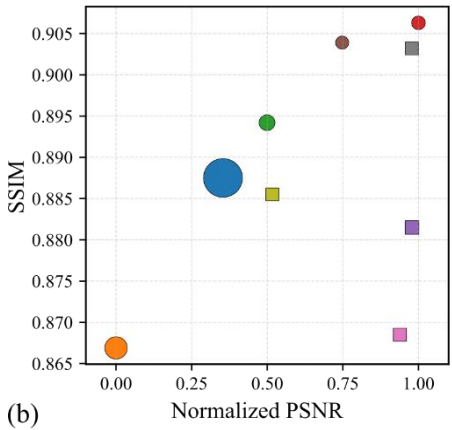

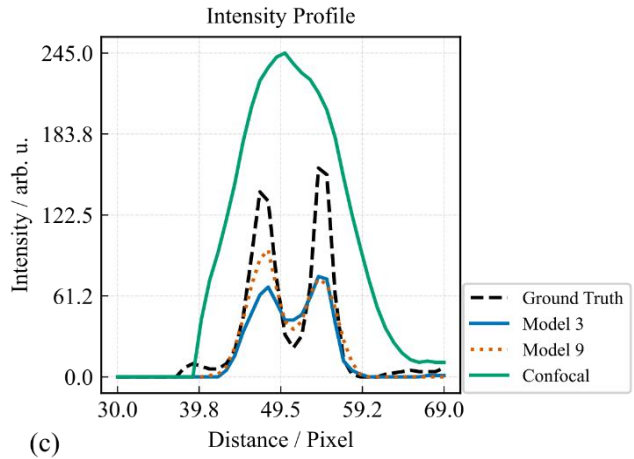

Figure 7. Quantitative and qualitative evaluation of nine CycleGAN models for confocal-to-deconvolved STED image reconstruction.
(a) Representative reconstruction results for a single confocal input image. The first image in the top row shows the ground truth dSTED image, followed by outputs from Models 1–9 in both the first and third rows. The second row shows, in order: zoomed-in crop of the ground truth, the confocal input, followed by a zoomed-in view and zoomed-in crops of the Model 1 and Model 7. This layout highlights differences in structural fidelity and signal localization. Yellow lines indicate the positions used for intensity profile analysis. PSNR and SSIM values for this sample, computed with respect to the ground truth, are reported in Table 6.

(b) Mean normalized PSNR and SSIM across the test dataset (Table 5). Each marker represents one model, with marker shape indicating the channel policy (circle: doubling; square: fixed) and marker size proportional to the number of generator parameters (Table 1).

(c) Intensity profiles plotted along the yellow lines in Figure 7a, comparing the confocal input, outputs from Model 3 and Model 9, and the ground truth dSTED image. Both models recover the two-peak structure observed in the ground truth, with Model 9 showing slightly closer alignment in peak intensity and separation. Model 3 reconstructs the overall shape but with lower peak contrast. The confocal profile remains broad and unresolved. These observations are supported by Pearson correlation coefficients with the ground truth profile: Model 9 (0.929), Model 3 (0.923), and confocal (0.731). All images feature the same spatial scale.

**Table 3. Average Image Quality Metrics Across Models for
Confocal to STED modality transfer (Full Dataset)**

| Model | 1 | 2 | 3 | 4 | 5 | 6 | 7 | 8 | 9 |
|---|---|---|---|---|---|---|---|---|---|
| SSIM | 0.887 | 0.905 | 0.903 | 0.901 | 0.899 | 0.878 | 0.904 | 0.910 | 0.909 |
| PSNR | 26.62 | 26.45 | 26.49 | 26.51 | 26.11 | 26.19 | 26.54 | 26.39 | 26.37 |

**Table 4. Image Quality Metrics for the Sample Shown in Figure 6a (Confocal to STED Transfer)**

| Model | 1 | 2 | 3 | 4 | 5 | 6 | 7 | 8 | 9 |
|---|---|---|---|---|---|---|---|---|---|
| SSIM | 0.95 | 0.93 | 0.94 | 0.94 | 0.95 | 0.93 | 0.95 | 0.96 | 0.98 |
| PSNR | 29.5 | 28.9 | 30.3 | 31.6 | 28.8 | 29.4 | 32.1 | 30.8 | 31.8 |

**Table 5. Average Image Quality Metrics Across Models for
Confocal to Deconvolved-STED modality transfer (Full Dataset)**

| Model | 1 | 2 | 3 | 4 | 5 | 6 | 7 | 8 | 9 |
|---|---|---|---|---|---|---|---|---|---|
| SSIM | 0.887 | 0.866 | 0.894 | 0.906 | 0.881 | 0.903 | 0.868 | 0.903 | 0.885 |
| PSNR | 26.62 | 26.25 | 26.77 | 27.29 | 27.26 | 27.02 | 27.22 | 27.26 | 26.79 |

**Table 6. Image Quality Metrics for the Sample Shown
in Figure 7a (Confocal to deconvolved-STED Transfer)**

| Model | 1 | 2 | 3 | 4 | 5 | 6 | 7 | 8 | 9 |
|---|---|---|---|---|---|---|---|---|---|
| SSIM | 0.94 | 0.94 | 0.94 | 0.93 | 0.94 | 0.94 | 0.92 | 0.92 | 0.94 |
| PSNR | 31.6 | 32.6 | 31.6 | 32.3 | 32.1 | 32.7 | 33.3 | 31 | 33.3 |

While lighter models, such as Model 9, significantly reduce computational demands, they can exhibit limitations such as blurring and spatial shifts in peak intensities. In contrast, larger models tend to retain finer structural detail and more accurate intensity profiles but may be prone to overfitting or amplifying noise. The architectural design thus presents a clear trade-off between model expressiveness and generalization capability, with fixed channel scaling emerging as an effective compromise that avoids excessive parameter growth while maintaining performance.

In addition to channel configuration, further reductions in model complexity can be achieved by implementing depth-wise separable convolutions, which decompose standard convolutions into lighter operations, and by reducing network depth, thereby limiting the number of layers and associated parameters. These strategies help retain image fidelity while minimizing memory footprint and computational cost.

Moreover, in GAN-based image translation, various hyperparameters and architectural design choices substantially influence model behavior and performance. These include the type of generator architecture, the formulation of adversarial and reconstruction losses, and the choice of normalization strategies. Selecting appropriate convolutional kernel sizes based on the spatial scale of the target structures and input image resolution is also critical for effective feature extraction. Training dynamics are further affected by factors such as learning rate scheduling, weight initialization schemes, and the use of regularization techniques like dropout. These elements collectively shape model stability, convergence, and generalization—particularly when adapting across datasets with differing noise characteristics and structural complexity. Lighter models make it substantially easier to test these variations systematically. Due to their reduced training time and memory consumption, such models enable rapid experimentation, hyperparameter optimization, and cross-validation across diverse datasets. Importantly, they also lower the computational burden, leading to reduced energy usage and significantly lower $CO_2$ emissions, contributing to more environmentally sustainable deep Learning Practices. As deep learning models continue to scale, these considerations become increasingly relevant—not only for technical optimization but also for responsible and resource-aware research.

*3.4. GAN-Based Inference as a Benchmark for Experimental and Labeling Fidelity*

In Figure 8a, we evaluate the utility of a GAN model—trained exclusively on high-quality STED data—to serve as a reference for assessing experimental integrity. The model was trained to transform confocal input images into their corresponding high-resolution STED representations. When applied to confocal data, the GAN-generated outputs were compared to experimentally acquired low-quality STED images, which have been compromised due to suboptimal imaging conditions, photobleaching, or inaccurate protein labeling. The discrepancy between the GAN output and the degraded STED data highlights the sensitivity of the model to subtle experimental deviations. In Figure 8b, the same approach was applied to a confocal-to-deconvolved transformation task. Here, the GAN was trained on high-quality deconvolved images and tested on

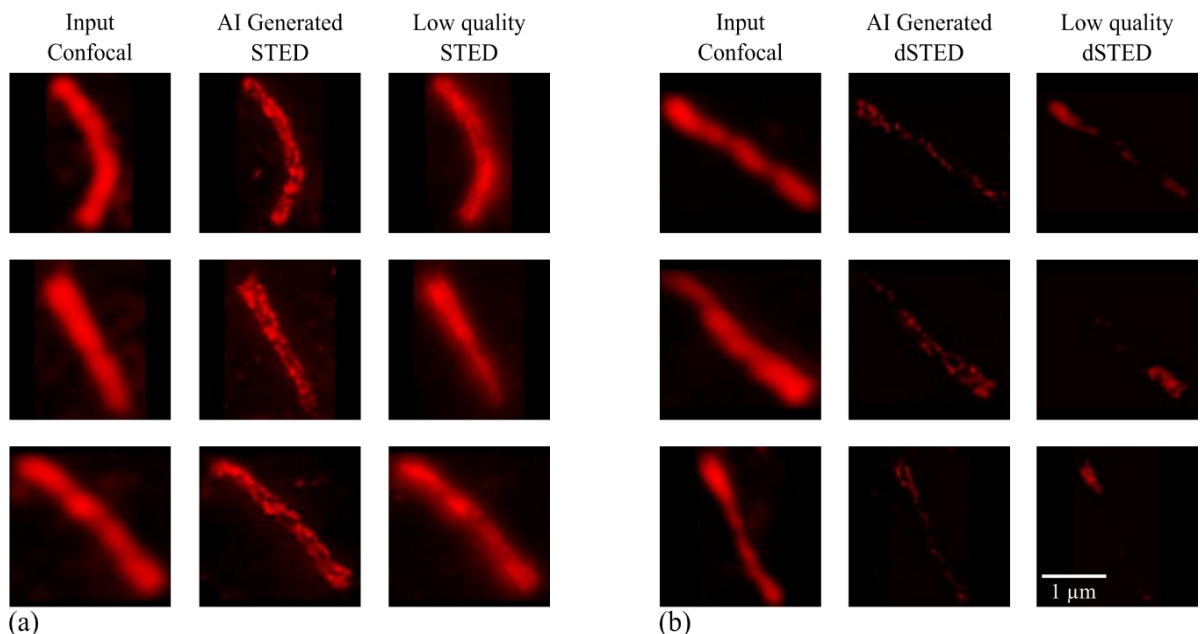

Figure 8. GAN-generated reconstructions trained on high-quality data can serve as a benchmark for evaluating the quality of experimental results.

(a) From right to left: confocal input, GAN-generated STED image trained on high-quality data, and an experimentally acquired low-quality STED image. The comparison illustrates how the GAN output deviates from the experimental image due to artifacts such as photobleaching or suboptimal labeling.

(b) From right to left: confocal input, GAN-generated deconvolved STED image trained on high-quality data, and a low-quality deconvolved STED image. The structural differences reflect the model's robustness and its potential to highlight experimental limitations when applied to non-ideal data. All images feature the same spatial scale.

corresponding confocal images. The results presented in Figure 8 demonstrate the potential of GAN-based models as benchmarks for evaluating experimental and labeling fidelity. A model trained on high-quality data captures an implicit standard of optimal structural features. Deviations between GAN predictions and new experimental outputs can indicate compromised imaging conditions or biological variability. This strategy is applicable across both STED and deconvolved STED modalities and enables indirect quality control in workflows where direct validation is limited. GAN-based inference thus offers a practical tool for identifying inconsistencies in sample preparation, imaging protocols, or labeling accuracy.

## 4. Conclusion:

In conclusion, this study presents a detailed comparison between CycleGAN and Pix2Pix for modality transfer tasks in fluorescence microscopy, specifically transforming confocal images into STED and deconvolved STED modalities. While Pix2Pix consistently achieved higher scores in SSIM and PSNR, CycleGAN demonstrated competitive performance. This is particularly relevant for the biomedical imaging domain, where generating paired datasets is often impractical due to the high cost, time, and complexity involved in biological sample preparation, precise labeling, and stable imaging conditions. Our findings suggest that CycleGAN remains a viable option when paired data is limited or unavailable.

We also investigated the influence of generator architecture by systematically reducing the number of channels in U-Net-based models. In doing so, we compared the traditional channel-doubling policy with a fixed channel strategy. The fixed channel models, particularly those with simplified architectures, achieved comparable or superior performance with significantly fewer parameters. This reduction not only improves training efficiency but also minimizes the risk of overfitting, making these architectures well-suited for low-data regimes and resource-constrained environments.

Beyond image translation, we introduced an application of GANs as tools for assessing experimental and labeling fidelity. By training models on high-quality data and comparing their outputs to low-quality experimental results, we showed that GAN predictions can reveal subtle deviations in imaging conditions or protein expression. This makes GAN-based inference a valuable diagnostic benchmark for validating experimental consistency, especially in high-throughput or large-scale microscopy studies. Together, these contributions highlight the utility of GANs not only for generating super-resolution images but also for guiding experimental reliability and promoting more reproducible imaging workflows.

## 5. Acknowledgment:

Christian Eggeling greatly acknowledge financial support for microscopy experiments by the Deutsche Forschungsgemeinschaft (DFG, German Research Foundation; Germany´s Excellence Strategy – EXC 2051 – Project-ID 390713860; project number 316213987 – SFB 1278; GRK M-M-M: GRK 2723/1 – 2023 – ID 44711651; GRK PhInt: GRK 3014/1; Instrument funding modular STED INST 1757/25-1 FUGG; project PolaRas EG 325/2-1), the State of Thuringia (TMWWDG), the Leibniz Association (Leibniz Collaborative Excellence Programme, project AMPel – project numer K548/2023), and the Free State of Thuringia (TAB; AdvancedSTED / FGZ: 2018 FGI 0022; Advanced Flu-Spec / 2020 FGZ: FGI 0031). Further, this work is supported by the BMFTR (Federal Ministry of Research, Technology and Space), Photonics Research Germany (FKZ: 13N15713 / 13N15717) and is integrated into the Leibniz Center for Photonics in Infection Research (LPI). The LPI initiated by Leibniz-IPHT, Leibniz-HKI, UKJ and FSU Jena is part of the BMFTR national roadmap for research infrastructures. A part of the project on which these results are based was funded by the Free State of Thuringia under the number 2018 IZN 0002 (Thimedop) and co-financed by funds from the European Union within the framework of the European Regional Development Fund (EFRE).